\documentclass{article}

\usepackage[english]{babel}

\usepackage[letterpaper,top=2cm,bottom=2cm,left=3cm,right=3cm,marginparwidth=1.75cm]{geometry}

\usepackage[affil-it]{authblk}

\usepackage{natbib}
\usepackage[utf8]{inputenc} 
\usepackage[T1]{fontenc}    
\usepackage[colorlinks, urlcolor=black,
citecolor=black, linkcolor=black]{hyperref}       
\usepackage{url}            
\usepackage{booktabs}       
\usepackage{amsfonts}       
\usepackage{nicefrac}       
\usepackage{microtype}      
\usepackage{xcolor}         

\usepackage{algorithm}
\usepackage{algpseudocode}
\usepackage{amsmath}
\usepackage{tikz-network}
\usepackage{pgfplots} 
\pgfplotsset{compat=newest}
\usepackage{graphicx}
\usepackage{caption}
\usepackage{subcaption}
\usepackage{comment}
\usepackage{wrapfig}

\title{Optimistic Optimization of Gaussian Process Samples}
\author[1,*]{Julia Grosse}\author[2]{Cheng Zhang} \author[1,3]{Philipp Hennig} 

\affil[1]{University of Tübingen}
\affil[2]{Microsoft Research Cambridge}
\affil[3]{Max Planck Institute for Intelligent Systems}
\affil[*]{Corresponding author: julia.grosse@uni-tuebingen.de}

\begin{document}
\maketitle

\begin{abstract}
     \emph{Bayesian optimization} is a popular formalism for global optimization, but its computational costs limit it to expensive-to-evaluate functions. A competing, \emph{computationally} more efficient, global optimization framework is \emph{optimistic optimization}, which exploits prior knowledge about the geometry of the search space in form of a dissimilarity function. We investigate to which degree the conceptual advantages of Bayesian Optimization can be combined with the computational efficiency of optimistic optimization. By mapping the kernel to a dissimilarity, we obtain an optimistic optimization algorithm for the Bayesian Optimization setting with a run-time of up to $\mathcal{O}(N \log N)$. As a high-level take-away we find that, when using stationary kernels on objectives of relatively low evaluation cost, optimistic optimization can be strongly preferable over Bayesian optimization, while for strongly coupled and parametric models, good implementations of Bayesian optimization can perform much better, even at low evaluation cost. We argue that there is a new research domain between geometric and probabilistic search, i.e.~methods that run drastically faster than traditional Bayesian optimization, while retaining some of the crucial functionality of Bayesian optimization.
\end{abstract}

\section{Introduction}
Bayesian optimization (BO) \citep{shahriari2015taking} is a popular and successful framework for global optimization. The foundation of most BO frameworks is a Gaussian Process (GP) regression method, whose kernel encodes prior knowledge about the objective function. This GP regressor provides a posterior over the objective, a probabilistic surrogate on which one can then reason about the extremum, and its location. This can be done in a variety of ways, e.g.~by using the posterior to construct upper bounds (in GP-UCB \cite{srinivas2009gaussian}), to estimate expected improvement (\cite{movckus1975bayesian}) or information gain (\cite{hennig2012entropy}, \cite{wang2017max}).  

The computational cost of BO itself is significant, and arises from at least two sources: First, exact GP inference has cost $\mathcal{O}(N^3)$, where $N$ is the number of observed function values (``samples''). Second, finding the next evaluation location requires a \emph{continuous, numerical} optimization of the acquisition function. Since this utility function is generally multimodal, optimization should, at least in principle, be carried out on an increasingly fine grid, which contributes up to $\mathcal{O}(N^{2d})$ costs \citep{salgia2021domain} 
(a third source of cost is the construction of the acquisition function itself, but there some popular choices, like GP-UCB, for which this step is essentially a trivial sum of the mean and marginal standard-deviation constructed during inference, of negligible overhead).
If the cost of individual function evaluations is very high, then the overhead of BO is irrelevant. But there are scenarios in which the cost of BO \emph{is} a concern, namely when individual function evaluations are of intermediate cost, and (perhaps as a direct consequence), the total number $N$ of evaluations is sufficiently large to make the cubic cost of GP inference noticeable. An example are settings involving computer simulations runs, e.g. in robotics, biology, chemistry and human-computer interaction design. 

Considering this ``middle ground'' between sample and computational efficiency, we study a competing framework for global optimization,   \emph{optimistic optimization} (OO),  which has drastically lower computational overhead. OO does not require computing an explicit global posterior on the objective. However, OO can nevertheless leverage at least certain kinds of prior knowledge, captured in the form of a dissimilarity measure that can be constructed in the form of a probabilistic deviation inequality for samples from a GP. This in effect produces a map from a GP prior one might otherwise use in BO to a (much more time-efficient) OO model, so that prior knowledge encoded in the GP can be leveraged without the intermediate step of (cubically expensive) GP regression. For noiseless (or near-noiseless) observations, the resulting OO algorithm achieves $\mathcal{O}(N \log N)$ computational cost. Interestingly, \emph{some} forms of prior information can not be leveraged in this way, so there \emph{are} settings in which BO is indeed preferrable, even in terms of raw overall speed.

In summary, we provide a practical map between BO and OO for global optimization (Section \ref{sec:connection}), resulting in a hybrid method we call GP-OO. We then contribute a formal analysis of this method in terms of regret (Section \ref{sec:theory}), also pointing out limitations in the process. Experiments (Section \ref{sec:experiements}) corroborate that the proposed GP-OO can be significantly more time-efficient than BO in settings with low (but nevertheless realistic) function evaluation costs.

\section{Background}
\subsection{Bayesian Optimization (BO)}
We consider maximizing a function $f: \mathcal{X} \rightarrow \mathbb{R}$. Throughout, $f^*$ denotes the maximum of the function, and $x^*$, the point where it is attained. $k: \mathcal{X} \times \mathcal{X} \rightarrow \mathbb{R}$ is a kernel function. The function $f$ is assumed to be a sample from a GP $\mathcal{GP}(\mu,k)$. We assume that the GP is centered, i.e. $\mu = 0$. One has access to observations $y_i$, where $y_i \sim f(x_i)+\mathcal{N}(0,\lambda)$. The noiseless setting amounts to the assumption $\lambda \rightarrow 0$. 
After each new observation the posterior over the function is updated:
\begin{align}
    \mu_{n}(x) = \mathbf{k}_n(x)^T(\mathbf{K}_n + \lambda I)^{-1} \mathbf{y}_n \hspace{1cm}
    k_n(x,x) = k(x,x) - \mathbf{k}_n(x)^T(\mathbf{K}_n + \lambda I)^{-1}\mathbf{k}_n(x),
\end{align}
where $\mathbf{y}_n = [y_1,..., y_n]^T,  \mathbf{k}_{n}(x) = [k(x_1,), ...,k(x_n,)]^T$, and $\mathbf{K}_n$ is the Gram matrix with $K_n{_{ij}} = k_n(x_i, x_j)$.
The posterior is used to define an acquisition function $a_n(x)$ at whose maximum the function is evaluated next. For an overview of different options of acquisition functions see \cite{frazier2018tutorial}. We will focus our analysis on the choice of GP-UCB \cite{srinivas2009gaussian}, where 
\[a_n(x) = \mu_{n}(x) + \beta_n ^{1/2} k_n(x,x)\quad \text{ and }\quad \beta_n ^{1/2} \text{ an appropriate constant.}\]
An advantage of GP-UCB over other choices is its comparatively simple structure, which facilitates not just implementation but also analysis. While other aquisition functions may well be more sample-efficient in concrete settings, this will not be important for what is to follow, because we are primarily concerned with run-time complexity, where GP-UCB is arguably indeed the fastest possible choice.
\begin{wrapfigure}[21]{L}{5cm}
\centering
    \includegraphics[scale=0.6]{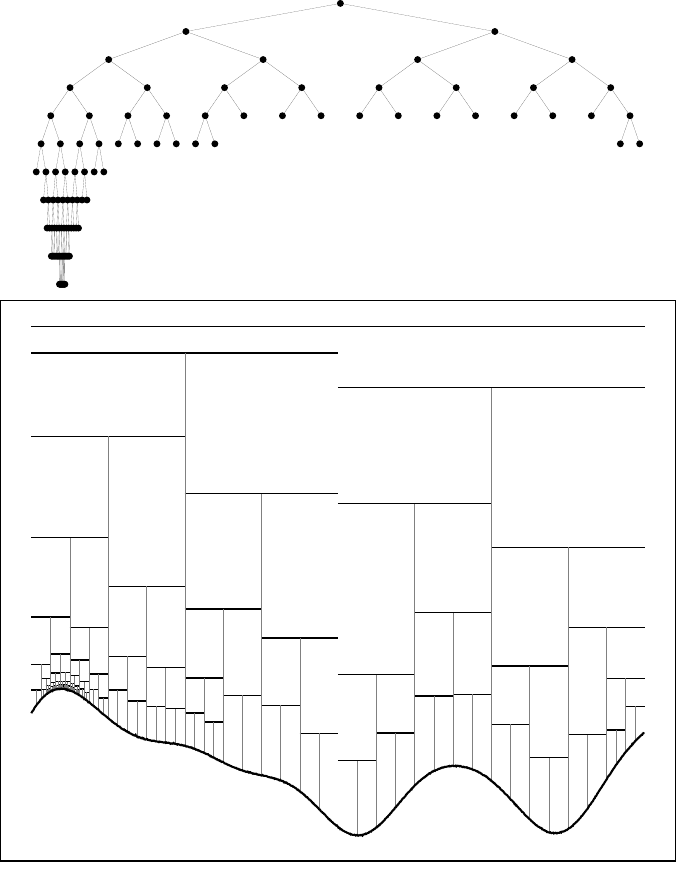}
    \caption{Optimistic optimization applied to a sample from a GP. The upper bounds are shown as horizontal bars. The vertical lines point to the evaluated locations.}
    \label{fig:optimisticoptimization}
\end{wrapfigure}

\subsection{Optimistic Optimization (OO)}
The optimistic optimization principle is used to optimize functions that are known to fulfill a local Lipschitz assumption with respect to a dissimilarity $d: \mathcal{X} \times \mathcal{X} \rightarrow \mathbb{R}$:
\begin{align}
\forall x \in \mathcal{X}: \, |f(x^*) - f(x)| \leq  d(x^*,x).
\end{align}
The method revolves around a hierarchical partitioning of the search space $\mathcal{X}$ that can be described by an infinite binary tree. Search can thus be implemented as a fast descent through the tree, thus only involves evaluations on a countable set of mesh points, in contrast to the local numerical optimization done within Bayesian optimization. This is one of the two reasons for the drastically shorter run-time costs of OO over BO (the other being that it does not require computing a GP posterior).

The root node corresponds to the entire search space $\mathcal{X}_{(0,1)}=\mathcal{X}$ and is named $(0,1)$. Consider a node $(t,i)$ at depth $t$. The left child $(t+1, 2i-1)$ and right child $(t+1,2i)$ represent two subregions $\mathcal{X}_{(t+1, 2i-1)}$ and $\mathcal{X}_{(t+1, 2i)}$ such that $\mathcal{X}_{(t,i)} = \mathcal{X}_{(t+1, 2i-1)} \cup \mathcal{X}_{(t+1, 2i)}$, i.e. the tree covers the entire space. To indicate that a cell was explored at iteration $n$, we refer to it as $\mathcal{X}_n$.
Intuitively, it makes sense to select the cells in such a way that all points in a cell are similar to each other and all similar points are in the same cell. Formally, this can be expressed as
\begin{itemize}
  \item[(a)] $\forall x,y \in \mathcal{X}_{(t,i)}: \{d(x,y) < \delta(t)\}$
  \item[(b)] $\exists x_{(t,i)} \in \mathcal{X}_{(t,i)}: \{y \in \mathcal{X}: d(x_{(t,i)},y) < c \, \delta(t)\} \subset \mathcal{X}_{(t,i)}$,
\end{itemize}
where $\delta(t)$ is a decreasing sequence of diameters and $c$ is a global constant.
During search, the tree is build incrementally by adding the two children of a selected node. When a new node ($t,i$) is added, an observation is made at the center $x_{(t,i)}$ of the region $\mathcal{X}_{(t,i)}$. In each round, the leaf with the highest upper bound $U_{(t,i)} = f(x_{(t,i)}) + \delta(t) $ is selected for expansion. Since $f$ is assumed to be locally Lipschitz with respect to $d$, selecting nodes by $U_{(t,i)}$ is a valid upper bound strategy (assuming noiseless observations). 
The sequence of $\delta(t)$ controls exploration. Therefore, the partitioning should be chosen in such a way, that the $\delta(t)$ can be as small as possible. The optimistic optimization principle is summarized in Algorithm 1 and Figure \ref{fig:optimisticoptimization}. Using a binary heap, the priority queue for the leaf nodes can be realized in $\mathcal{O}(N \log N)$ if the budget $N$ is known in advance.
\begin{algorithm}
\begin{algorithmic}[1]
\Procedure{Optimistic optimization}{$f, \delta$} 
\State priority-queue $\gets$ [(0,1)] //root node
\For{$j=1$ to $n$} 
    \State Select node $(t,i)$ with maximum $U_{(t,i)} = f(x_{(t,i)})+ \delta(t)$ from the priority-queue
    \State Observe $f(x_{(t+1,2i-1)})$ and $f(x_{(t+1,2i)})$ for the two children of $(t,i)$.
    \State Calculate child utilities $U_{(t+1,2i-1)} = f(x_{(t+1,2i-1)}) + \delta(t+1)$ and $U_{(t+1,2i)} = f(x_{(t+1,2i)}) +\delta(t+1)$.
    \State Add children $({t+1,2i-1})$ and $({t+1,2i})$ to the priority-queue based on their utilities.
\EndFor 
\EndProcedure
\end{algorithmic}
\caption{The optimistic optimization principle.}
\end{algorithm}

This algorithm is a batch method in the sense that all children of a newly explored node are evaluated in each step. We assume batch size two, but other choices are possible too and might be preferable in some applications, e.g.~when parallel evaluation is possible. A conceptual difference to the upper bound in GP-UCB is that one here works with upper bounds for entire \emph{regions} of the search space, not for individual points in it. Also, the exploration term $\delta(t)$ in the upper bound is not updated based on new observations, but derived a priori. Nevertheless, the exploration term, aka.~the uncertainty decreases during the search as the tree grows. 
 
The hierarchical optimistic optimization principle has its origins in the Bandit setting. \cite{bubeck2011x} apply it in the noisy setting, \cite{munos2011optimistic} apply it in the noiseless setting and \cite{kleinberg2008multi} uses it with a slightly different Lipschitz assumption. The assumptions on the dissimilarity $d$ vary, e.g. some theoretical analyses require that it additionally is a semi-metric, or metric.
\cite{munos2011optimistic}  introduces a variant of the principle, simultaneous optimistic optimization (SOO), for situations in which the dissimilarity function is unknown.  \cite{valko2013stochastic} extend this idea to the noisy setting.

\textbf{An interim summary:} OO descends along a search tree, i.e.~in a \emph{discrete} sequence of steps, without numerical optimization, and only using \emph{local} summary statistics, rather than updating a global posterior.
This makes OO very fast, at least compared to Bayesian optimization. But the approach also has a downside: At least at first sight, it is not clear how to encode salient prior information about the global optimization problem into the algorithm. By contrast, many Bayesian optimization experts see the rich language of GP prior models as a key strength of their framework. The following section thus investigates formal connections between OO and BO. The goal is to understand to which degree the structural language of a GP prior can be translated into the algorithmic efficiency of OO.
\section{Connection between Bayesian and Optimistic Optimization}
\label{sec:connection}

The policy of OO is based on measuring similarity in the input domain in terms of a metric defined through function values. It turns out that Gaussian process models -- or, more precisely, kernels -- can be used immediately to define such a pseudo-metric (Section 3.1). We further show how the pseudo-metric can be used to obtain upper bounds on the supremum of the cell (Section 3.2), that allow for the application of the OO principle in the BO setting. Section 3.3 is concerned with how to choose the cells and in Section 3.4 we illustrate the derived concept (GP-OO) on concrete examples. 
\subsection{Canonical pseudo-metric}
The canonical pseudo-metric $d: \mathcal{X} \times\mathcal{X} \rightarrow \mathbb{R}$ for a centered GP is defined as
\begin{align*}
d(x,y) = \mathbb{E}_{f \sim \mathcal{GP}(0,k)}[(f(x)-f(y))^2]
= \sqrt{k(x,x)+k(y,y)-2k(x,y)}.
\end{align*}
The fundamental relevancy of the canonical pseudo-metric arises from the fact that it controls the increments of the GP, in the sense of the following deviation inequality (\cite{pisier1999volume}, Theorem 4.7):
\begin{align*}
\forall u > 0, \mathbb{P}(|f(x)-f(y)| \geq u) \leq 2 \exp \biggl(- \frac{u^2}{2d(x,y)^2} \biggr)
\end{align*}
An important class of kernel-induced metrics is formed by monotonic transformations of the Euclidean metric, i.e. $d(x,y) = g(\|x-y\|_2)$ where $g: \mathbb{R} \mapsto \mathbb{R}$ is monotonically increasing. Many kernels used in practice are in this class, e.g. the square-exponential kernel, the Mat\'{e}rn class of kernels, the rational-quadratic kernels and the Wiener kernel as well as sums and products thereof. We refer to this class of kernels as $\mathcal{K}$. Kernels \emph{violating} this assumptions are e.g. polynomial or periodic kernels. 
\subsection{Upper bound on the supremum of a cell}
The first challenge in applying the optimistic optimization principle on samples of a GP consists in the probabilistic nature of the deviation inequality. To obtain a valid upper bound for the maximal deviation in a cell $\sup_{x \in \mathcal{X}_n} |f(x)-f(x_{n})|$, the deviation for \textit{all} points in the cell has to be bounded.  We approximate such an upper bound by introducing two simplifications: We discretize the search space $\mathcal{X}$ into a finite number of points $\hat{\mathcal{X}}$ and we introduce an independence assumption between the $|f(x)-f(x_{n})|$. Then we take a union bound approach:
\begin{align}
\mathbb{P}&(\sup_{x \in \hat{\mathcal{X}}_{n}} |f(x)-f(x_{n})| \geq u)
 &&\leq \sum_{x \in \hat{\mathcal{X}}_{n}} \mathbb{P}(|f(x)-f(x_{n})|\geq u) \\
 &\leq 2 \sum_{x \in \hat{\mathcal{X}}_{n}} \exp \biggl(- \frac{u^2}{2d(x,y)^2} \biggr)
 &&\leq 2 |\hat{\mathcal{X}}_{n}| \exp(-\frac{u^2}{2\Delta(\mathcal{X}_n)^2})\\
 \label{eq:Delta}
\text{where } \Delta(\mathcal{X}_n) &= \max_{x \in \mathcal{X}_{n}} d(x_{n}, x).
\end{align}
The bounds have to hold at each step $n$, so we additionally take a union bound over the number of steps. This implies the following statement, that holds with high probability :
\begin{align}
    \forall n: \sup_{x \in  \hat{\mathcal{X}}_{n}}|f(x)-f(x_{n})| \leq \beta_n^{1/2} \Delta(\mathcal{X}_n)
\end{align} 
where $\beta_n$ are appropriate constants specified in the Supplements.
The union bound approximation will be good if the $|f(x)-f(x_{n})|$ are (nearly) uncorrelated, or the size of the discretization $|\hat{\mathcal{X}}|$ is chosen sufficiently small. Otherwise the bounds are loose, which leads to over-exploration. Though approximate, this idea of neglecting correlations to simplify the calculation of the expected supremum of dependent Gaussian variables is, for example, also done in Maximum Value Entropy Search \citep{wang2017max} for BO or in related settings \citep{grosse2021probabilistic}. The other extreme, a greedy approach with $\beta_n = 1$ has also been taken in recent work \citep{rando2021ada}. 
With generic chaining \citep{talagrand1996majorizing} it is possible to improve over the union bound approach. However, to the best of our knowledge state-of-the-art algorithms \citep{borst2020majorizing} to optimize for tighter bounds require polynomial run-time in the number of points per cell for arbitrary kernel functions. For special cases, like a Wiener kernel \citep{talagrand2021empirical} or Mat\'{e}rn kernel functions \citep{shekhar2018gaussian} analytical attempts to derive chaining based upper bounds exist.
\subsection{Choosing the partitioning}
The second main step in applying the OO principle is to choose the cells and location of the centers in such a way that the diameters of the cells are as small as possible. 
For $k$ children nodes, this is a metric $k$-center problem -- one of the classical NP-hard problems \citep{gonzalez1985clustering}. A greedy approximation consists in iteratively picking the $k$ centers with the largest distance to the previously picked centers, and requires $\mathcal{O}(|\hat{\mathcal{X}}|k)$ time. The greedy procedure is guaranteed to result in a $2$-approximation, and there is no polynomial time algorithm doing better (unless P$=$NP). Working with a greedy instead of the optimal partitioning scheme thereby leads to an additional factor of $2$ in the below regret bound, but is not harmful in the sense that the search gets stuck in a local optimum. NP-hardness also appears in the context of BO, e.g. the exploration term used in GP-UCB. And the acquisition functions of information-theoretic BO methods \citep{hennig2012entropy}, \citep{wang2017max} are related to the maximization of information gain, which is also a NP-hard problem. \\
From an implementation perspective, it is desirable to constrain the partitioning to axis-parallel boxes. For some kernel functions, e.g. the polynomial kernel, requirement (b) of the OO principle cannot be fulfilled with axis parallel boxes. One can nevertheless run the algorithm, but it will clearly be less information-efficient.
One may even consider a randomized choice of centers, e.g. as done in Monte Carlo Tree Search \citep{chaslot2008monte}. However, it still remains to calculate the maximal distance from a point in the cell $\mathcal{X}$ to the center $x_n$. 
The computational complexity of this part is comparable to the numerical optimization of the acquisition function in BO. An advantage is that the domain over which one optimizes shrinks in each step. A disadvantage, though, is that if the numerical optimization is suboptimal and the maximal distance within a cell is underestimated, cells containing the optimum might get irreversibly pruned. \\
An important observation is that the partitioning scheme itself does not depend on the objective $f$, but only the kernel/distance function (how the search tree grows, however, \emph{does} depend on $f$). This opens up the possibility of finding a good partitioning and the corresponding maximal distances analytically. For distances derived from kernels in $\mathcal{K}$, one can apply the following \textit{regular} partitioning scheme: At each step, cut along the longest dimension in order to obtain the two children cells. Use the euclidean centers as centers. A point that maximizes the distance to the center will always be one of the $2^m$ corner points in $m$ dimensions. Thus, for this type of kernels, the costs reduce to $\mathcal{O}(N2^m)$ for the partitioning, or $\mathcal{O}(N \log N + N 2^m) = \mathcal{O}(N \log N)$ in total. 

\begin{figure}[t]
    \centering
    \includegraphics{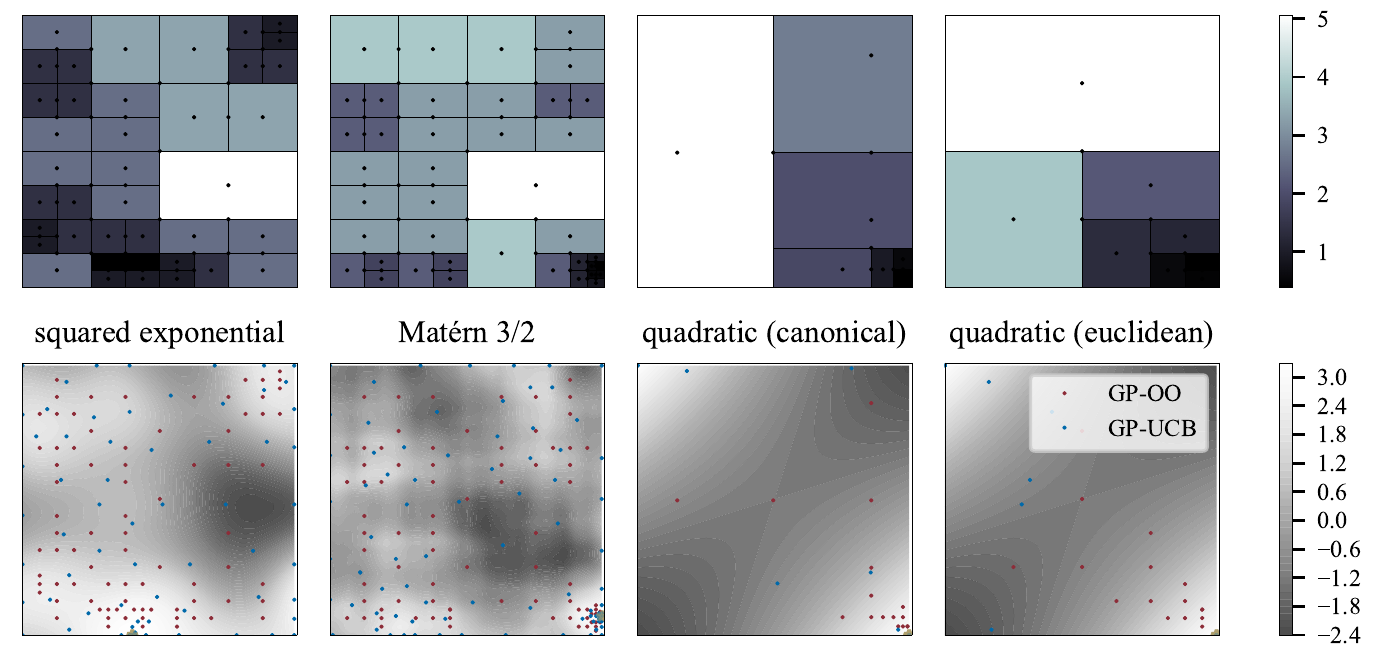}
    \caption{Top row: Cells and diameters $\Delta$. Brighter colors indicate larger diameters. Bottom row: Sample from a GP and evaluation locations of GP-OO and GP-UCB.}
    \label{fig:partitions_and_contours}
\end{figure}

\subsection{GP-OO}
Motivated by the analysis above, we propose a variant of OO, which we call GP-OO. It consists in running Algorithm 1 with the utility $\mathcal{U}_{(n)} = f(x_{(n)})+ \beta_n^{1/2} \Delta(\mathcal{X}_n)$ in line 4, where $\Delta$ is as defined in Eq.~\eqref{eq:Delta}. For kernels that match the observation above, one can then still choose the evaluation node (in $\mathcal{O}(2^m)$ time) without numerical optimization. Figure \ref{fig:partitions_and_contours} shows the algorithm running on GP samples from a square-exponential and a Mat\'{e}rn kernel with $\nu= 2/3$ on the domain $[0,5]^2$, as well as from a quadratic kernel on the domain $[-1, 1]^2$. In regions with higher function values the partitions are finer. The Mat\'{e}rn kernel yields higher distances than the square exponential, leading to more exploration, reflecting that the samples are less smooth. By optimizing the partitions and centers of the cells with respect to the canonical pseudo-metric, larger parts of the search space can be covered while keeping the cell diameters constant as shown by the two examples with the quadratic kernel. 

\section{Related Work}
\begin{figure}[t]
\centering
    \centering
    \includegraphics[width=\textwidth]{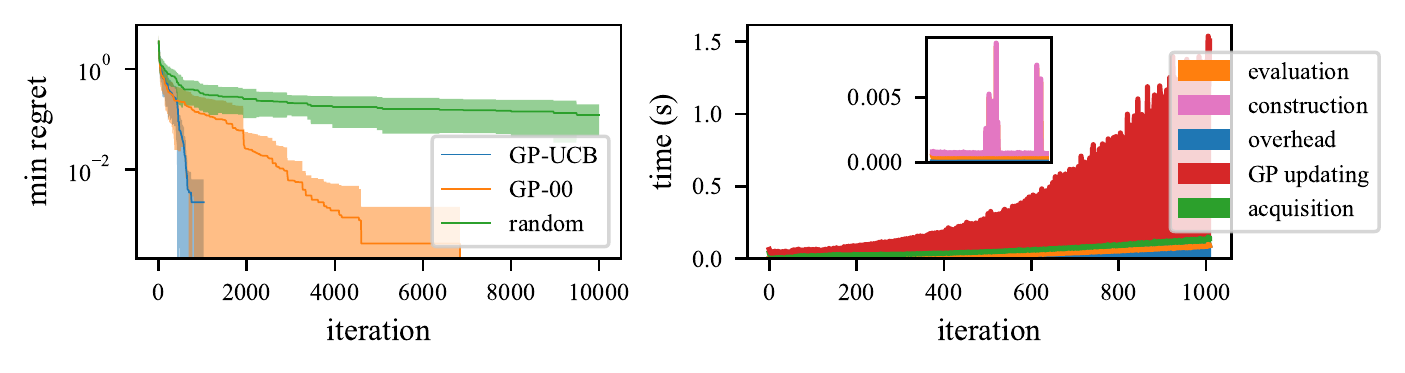}
    \caption{Left: Minimum simple regret $|f(x^*)-f(x)|$ on synthetic samples. Right: Composition of the average computational costs in GP-UCB (large panel) and GP-OO (small panel, note that its ordinate is re-scaled by a factor of $10^3$).}
    \label{fig:compositionofcosts}
\end{figure}

In this section we review related work at the intersection of OO and BO. 
With the exception of \cite{grill2018optimistic}, the fundamental difference to all of this work is that we do not keep track of a GP-posterior, thus saving significant computational cost (see Figure \ref{fig:compositionofcosts}).\\
\textbf{Work without the canonical pseudo-metric.} BO methods have been combined with SOO (\cite{munos2011optimistic}), the version of OO with unknown dissimilarity. In BamSOO, \cite{wang2014bayesian} use GP-UCB to reduce the number of evaluations required when running SOO alone. By using SOO, they can in return reduce the optimization costs of the acquisition function.
 \cite{gupta2021bayesian} improve upon the basic version of BamSOO by a more elaborate partitioning scheme: Instead of dividing a cell into $k$ children along the longest side of the cell, they divide along the $b$ longest dimensions into $a$ cells, where $b^{a}=k$. 
 \cite{salgia2021domain} use a random walk based strategy on a tree to improve over the grid-based optimization of the GP-UCB acquisition function. 
 For the Mat\'{e}rn and Squared Exponential kernel, they achieve order optimal regret, but the computational complexity is $\mathcal{O}(N^4)$.\\
\textbf{Work with the canonical pseudo-metric.}
\cite{shekhar2018gaussian} use the GP's canonical pseudo-metric to improve the numerical optimization by pruning the search regions. They additionally keep track of the posterior to only evaluate at locations where posterior uncertainty exceeds the cell's upper bound. 
\cite{rando2021ada} follow this approach and additionally introduce a Nyström approximation, which allows for approximate inference in $\mathcal{O}(N^2d_{\text{eff}}^2)$, where $d_{\text{eff}}$ is the effective dimension of the search space. 
\cite{contal2015optimization} replace the GP-UCB bounds with bounds derived from the pseudo-metric, but do not use a hierarchical approach, i.e. they construct bounds for individual points. They update the posterior and the exploration terms after every new observation. \\
\cite{grill2018optimistic} apply the optimistic optimization principle to a one-dimensional Brownian walk. A minor difference is, that they evaluate a cell at the corners of an interval and not in the center. There are cases, where this is advantageous, e.g. think of samples from a GP with a polynomial or linear kernel. However, the number of corners scales exponentially with the dimension.
\section{Regret}
\label{sec:theory}
While computational and not sample efficiency is our main motivation to apply the OO principle in the BO setting, we show that the resulting method nevertheless leads to non-trivial regret. In particular, it is  asymptotically regret-free in the limit $\lim_{N \rightarrow \infty} R_N/N$. Here, $R_N$ denotes the cumulative regret defined as $R_N = \sum_{n=0}^N f(x^*) - f(x_n)$.\\
Building upon arguments from \cite{munos2011optimistic} and \cite{shekhar2018gaussian}, we obtain the following guarantee for the cumulative regret:\\
\textbf{Proposition 1}
\textit{Let $\mathcal{X}$ be finite, $\epsilon \in (0,1)$ and $\beta_n = 2 \log(N|\mathcal{X}_n|/2\epsilon)$. Running GP-OO with $\beta_n$ for a sample $f$ from a GP with mean function zero and covariance function $k(x,x)$, the following regret bound holds with probability $1-\epsilon$:}
\begin{align*}
R_N \leq \beta^{1/2} \sum_{n=1}^N \Delta(\log n)
\end{align*}
\textit{where $\beta = max\{\beta_1,..., \beta_N\}$ and $\Delta(n)$ denotes the diameter of a cell evaluated at depth $n$.}

A full proof is provided in the Supplements. The high-level idea is that either the explored cell $\mathcal{X}_n$ contains the optimum $x^*$, then the simple regret $|f(x^*)-f(x)|$ is trivially upper bounded by the maximal deviation $\beta^{1/2}\Delta(\mathcal{X}_n)$ . Or the cell does not contain the optimum, but then then its utility was higher than the one of a node containing the optimum in its region, and thereby higher than the optimum itself. For the cumulative regret we assume the worst-case of a uniformly growing tree.
For a broad class of kernels, the  bound in Proposition 1 can be further specified:\\
\textbf{Proposition 2}
\textit{Assume the GP's canonical pseudo-metric $d$ satisfies $d(x,y) = C\|x-y\|_2^{\alpha}$, where $C > 0, \alpha > 0, m / \alpha > 1$. Running GP-OO on a finite domain $ \mathcal{X} \subset [0,1]^m$ with regular partitions, one has a worst-case cumulative regret $R_N$ of}
\begin{align*}
\tilde{\mathcal{O}}(N^{1-\alpha/m}(\log N)^{\alpha/m})
\end{align*}
\textit{with high probability The $\tilde{\mathcal{O}}(\cdot)$ notation supresses poly-logarithmic factors}.\\
In particular, for the squared exponential kernel and Mat\'{e}rn kernels with half-integer values $\nu \geq 3/2$, one has $\alpha=1$ and $R_N \in \tilde{\mathcal{O}}(N^{1-1/m}(\log N)^{1/m})$.
 For comparison, the regret in GP-UCB grows as $\tilde{\mathcal{O}}(\sqrt{N} \log(N)^{\frac{m+1}{2}})$ for the squared exponential kernels and thereby scales better to higher dimensions. For the Mat\'{e}rn class, GP-UCB is guaranteed to have regret at most $\tilde{\mathcal{O}}(N^{\frac{\nu+m(m+1)}{2\nu+m(m+1)}})$ for $\nu \geq 1$. Our bound is tighter, e.g., for the values $\nu = 3/2$ or $\nu = 5/2$ often used in practice. 

 \cite{shekhar2018gaussian} establish regret bounds in terms of the near-optimality dimension $\tilde{m}$. This measure is commonly used in optimistic optimization to characterize the size of the set of $\epsilon$-optimal points in terms of packing numbers. The near-optimality dimension does not only depend on the underlying metric, but also on the function $f$ itself and is thereby a random variable. Smaller values are associated with deeper growing trees, whereas larger values lead to more balanced, uniformly growing trees. Assuming the worst case of $\tilde{m}=m$, the bounds in  \cite{shekhar2018gaussian} become $\tilde{\mathcal{O}}(N^{1-\alpha/m})$ for the Mat\'{e}rn class and match our worst-case bound. However, we restricted the analysis to finite domains $\mathcal{X}$, and $|\mathcal{X}|$ enters our bound logarithmically. 
 
 \cite{grill2018optimistic} showed that in the specific case of Brownian motion, the tree built during optimistic optimization does not grow with the worst-case uniform rate.

\section{Experiments}
\label{sec:experiements}
We empirically compare GP-OO to GP-UCB in terms of regret and time, on synthetic three dimensional samples from Gaussian processes, and Benchmark functions from \cite{simulationlib}. All experiments were performed on a desktop machine (hardware details in the Supplements). Since computational efficiency is the key concern of our analysis, we often report performance relative to wall-clock time. We recognise, however, that such results are implementation dependent, and should be interpreted qualitatively. For GP-UCB, we use implementations from Emukit \citep{emukit2019, Gardner2018}.
\subsection{Experiments on synthetic functions}

\begin{figure}[t]
    \centering
    \includegraphics[width=\textwidth]{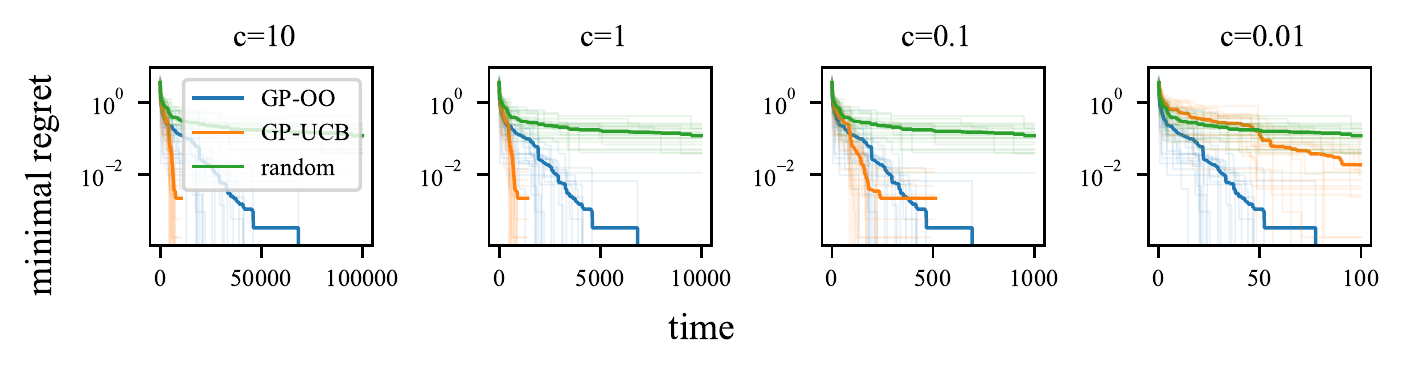}
    \caption{Minimum simple regret $|f(^*)-f(x)|$ on synthetic samples, for different function evaluation costs $c$, from high (10sec, left) to low (0.01sec, right). GP-UCB becomes less competitive as evaluation cost drops. For extremely cheap functions, purely random search eventually outperforms everything, because it can simply cover the whole region at nearly no overhead. Note the varying abscissas between plots. The line for GP-UCB changes nontrivially, reflecting its significant numerical overhead compared to the other two methods.
    Details in text.}
    \label{fig:functionevaluationcosts}
\end{figure}

\textbf{Regular partitions.} We begin with on-model experiments for a common setting in Bayesian optimization, where GP-OO can showcase its speed advantages: 20 samples from a GP with squared exponential kernel (lengthscale $l=0.1$) on the unit cube $\mathcal{X} = [0,1]^3$.  For all experiments, the noise level for GP-UCB is set to a very small constant $\lambda = 0.001$ since we assume noiseless observations. For GP-OO, we use $\beta_n = 2 \log(|\hat{\mathcal{X}_n}|N/2\epsilon|)$ with fixed horizon $N=10000$, and for GP-UCB, we use $\beta_n = 2 \log(|\hat{\mathcal{X}}|n^2\pi^2/6\epsilon)$ according to theory. We did a grid search over the size of the discretization $|\hat{\mathcal{X}}|$ with values in $\{1, 10, 100, 1000\}$. The confidence level $\epsilon$ is set to $0.01$ for both methods, and we set $|\hat{\mathcal{X}}|=1$. Even though GP-UCB is more sample-efficient than GP-OO (Figure \ref{fig:compositionofcosts}, left) GP-OO is orders of magnitude faster per step (Figure \ref{fig:compositionofcosts}, right).

This suggests there is a ``sweet-spot'': We artificially added $c \in \{0.01, 0.1, 1, 10\}$ seconds to the per-evaluation costs to simulate different evaluation costs (Figure \ref{fig:functionevaluationcosts}). For evaluation time around 0.1 seconds and below, GP-OO becomes faster than GP-UCB. When the objective becomes ``expensive'', BO can leverage its sample efficiency (at the very extreme end of nearly instant function-evaluations, random search, which has negligible overhead, is always a trivially optimal asymptotic baseline).

\begin{figure}[b]
    \centering
    \includegraphics[width=\textwidth]{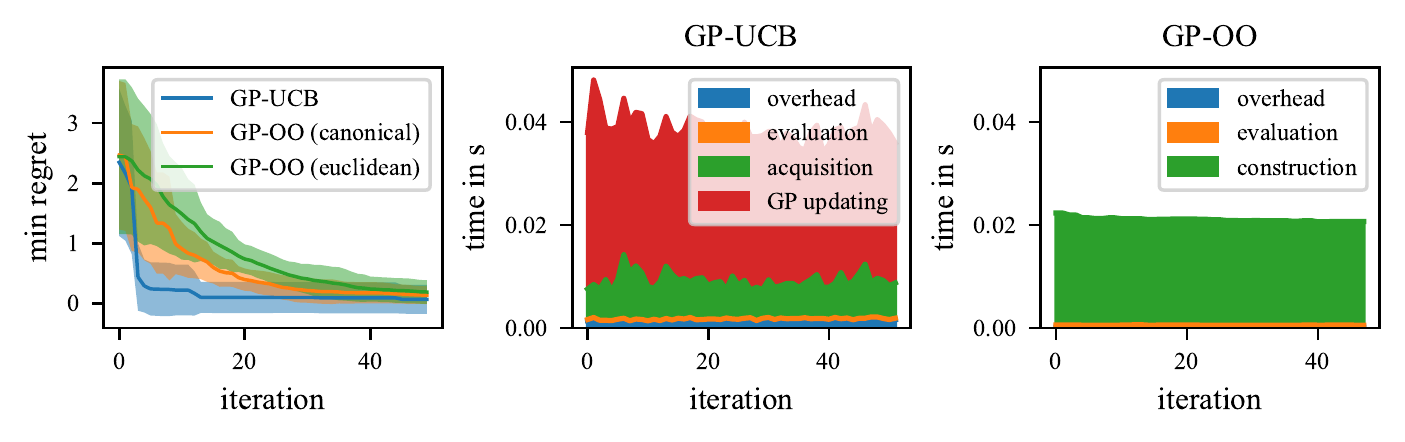}
    \caption{Experiment with a quadratic kernel, a problem specifically chosen such that GP-OO can not be expected to work well. Left: Performance in terms of the minimal obtained simple regret $|f(x^*)-f(x)|$. Middle: Wall-clock time with GP-UCB. Right: Wall-clock time with GP-OO.}
    \label{fig:polynomial_timing_area}
\end{figure}

\textbf{Non-regular partitions.} For demonstration, we also explore a setting with non-regular optimal partitions, where GP-OO can not perform as well. We sample 100 functions from a GP with quadratic kernel $k(x,y)=(x^Ty)^2$ with bias 0 on the domain $[-1,1]^3$. We consider GP-UCB and GP-OO, once with euclidean partitions and once with partitions optimized with respect to the canonical metric. The exploration constant was set to $\beta=1$. The results shown (Figure \ref{fig:polynomial_timing_area}) show that it is advantageous to optimize the partitioning scheme with respect to the canonical metric. However, we restrict ourselves to partitions with axis-aligned cells, i.e. the perfectly correlated corner points with $k(x,y)=1$ and $d(x,y)=0$ end up in different cells. At this point, GP-UCB has a clear advantage, as information can be shared across the search space between the corner points. Also, the run-time advantage decreases due to the numerical optimization of the partitions. 

\subsection{Experiments on benchmarks}
\begin{figure}
    \centering
    \includegraphics[width=\textwidth]{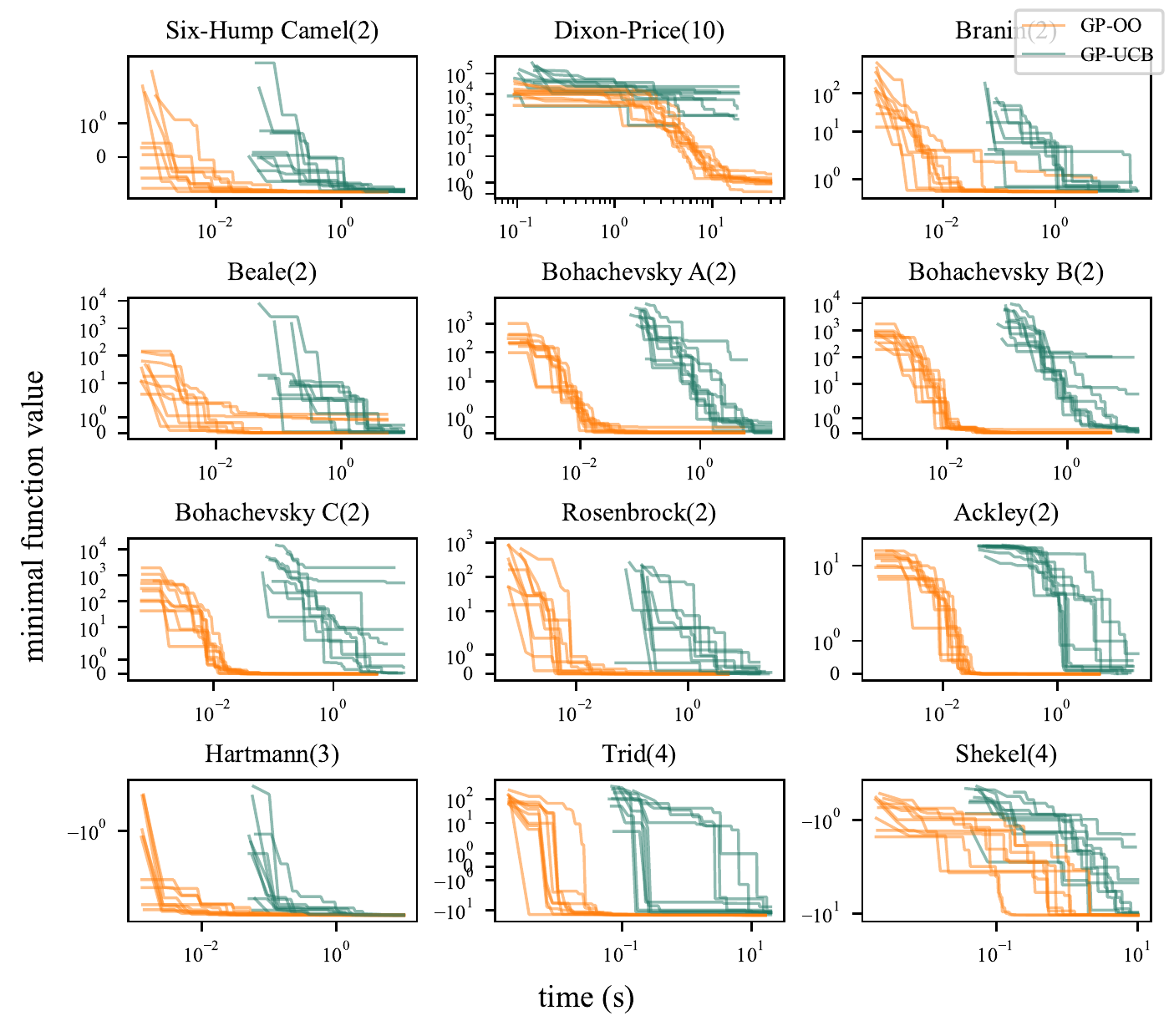}
    \caption{Optimization performance (minimal function values found) on common benchmarks with GP-UCB and GP-OO. The number in brackets indicates the dimension of the search domain. Note that all plots show wall-clock time in log-scale, not number of function evaluations, highlighting the strong \emph{computational} advantage of GP-OO over GP-UCB (as opposed to sample efficiency).}
    \label{fig:benchmark_times}
\end{figure}

We consider the minimization of common optimization benchmark functions, on domains with dimensions ranging from 2 to 10. Since the functions are deterministic, and GP-OO is deterministic up to random tie breaks, we randomly sample 10 domains $\mathcal{X}$ to obtain variation (see Supplements). We use a Mat\'{e}rn kernel with $\nu=3/2$ and length-scales as recommended by \cite{rando2021ada}. For the exploration constant $\beta$ we did a grid search over $\{0.1, 1, 10, 100\}$. We use confidence level $\epsilon = 0.05$.  Figure \ref{fig:benchmark_times} shows minimal attained function value over time. Here, GP-OO outperforms GP-UCB in many cases. This is not just due to runtime, but in fact also sometimes holds in function evaluations (see Supplements). Here, both methods did not always find the global optimum (which can not be guaranteed here as these functions are not necessarily within the RKHS).

\section{Conclusion}
BO and OO are closely connected through a mapping from the kernel-function to the canonical pseudo-metric of a GP. We showed that, for some, though certainly not all kernels, this connection can be exploited to derive a computationally efficient method for the BO setting that captures prior information. 
For common choices of kernels, like the Mat\'{e}rn class, we outperform even computationally light-weight BO approaches, like GP-UCB, if the objective function has reasonably low evaluation cost. Where the objective function is truly \emph{very} expensive, BO remains competitive. And, for strongly coupled models, too, we find BO to be advantageous. 
Our work shows that the ability to use prior information in Bayesian optimization does not \emph{have} to be expensive per se, but can also be achieved without explicitly tracking a posterior.

\section{Acknowledgements}
This work was supported by Microsoft Research through its PhD Scholarship Programme. The authors thank the International Max Planck Research School for Intelligent Systems (IMPRS-IS) for supporting Julia Grosse by non-financial means. The authors gratefully acknowledge financial support by the European Research through ERC StG Action 757275 / PANAMA; the DFG Cluster of Excellence "Machine Learning - New Perspectives for Science", EXC 2064/1, project number 390727645; the German Federal Ministry of Education and Research (BMBF) through the Tübingen AI Center (FKZ: 01IS18039A); and funds from the Ministry of Science, Research and Arts of the State of Baden-Württemberg.
\newpage
\bibliographystyle{alpha}
\bibliography{sample}

\begin{thebibliography}{27}
\providecommand{\natexlab}[1]{#1}
\providecommand{\url}[1]{\texttt{#1}}
\expandafter\ifx\csname urlstyle\endcsname\relax
  \providecommand{\doi}[1]{doi: #1}\else
  \providecommand{\doi}{doi: \begingroup \urlstyle{rm}\Url}\fi

\bibitem[Shahriari et~al.(2015)Shahriari, Swersky, Wang, Adams, and
  De~Freitas]{shahriari2015taking}
Bobak Shahriari, Kevin Swersky, Ziyu Wang, Ryan~P Adams, and Nando De~Freitas.
\newblock Taking the human out of the loop: A review of {B}ayesian
  optimization.
\newblock \emph{Proceedings of the IEEE}, 104\penalty0 (1):\penalty0 148--175,
  2015.

\bibitem[Srinivas et~al.(2009)Srinivas, Krause, Kakade, and
  Seeger]{srinivas2009gaussian}
Niranjan Srinivas, Andreas Krause, Sham~M Kakade, and Matthias Seeger.
\newblock {G}aussian process optimization in the bandit setting: No regret and
  experimental design.
\newblock \emph{arXiv preprint arXiv:0912.3995}, 2009.

\bibitem[Mo{\v{c}}kus(1975)]{movckus1975bayesian}
Jonas Mo{\v{c}}kus.
\newblock On {B}ayesian methods for seeking the extremum.
\newblock In \emph{Optimization techniques IFIP technical conference}, pages
  400--404. Springer, 1975.

\bibitem[Hennig and Schuler(2012)]{hennig2012entropy}
Philipp Hennig and Christian~J Schuler.
\newblock Entropy {S}earch for {I}nformation-{E}fficient {G}lobal
  {O}ptimization.
\newblock \emph{Journal of Machine Learning Research}, 13\penalty0 (6), 2012.

\bibitem[Wang and Jegelka(2017)]{wang2017max}
Zi~Wang and Stefanie Jegelka.
\newblock Max-value entropy search for efficient {B}ayesian optimization.
\newblock In \emph{International Conference on Machine Learning}, pages
  3627--3635. PMLR, 2017.

\bibitem[Salgia et~al.(2021)Salgia, Vakili, and Zhao]{salgia2021domain}
Sudeep Salgia, Sattar Vakili, and Qing Zhao.
\newblock {A} {D}omain-{S}hrinking based {B}ayesian {O}ptimization {A}lgorithm
  with {O}rder-{O}ptimal {R}egret {P}erformance.
\newblock In \emph{Thirty-Fifth Conference on Neural Information Processing
  Systems}, 2021.

\bibitem[Frazier(2018)]{frazier2018tutorial}
Peter~I Frazier.
\newblock A tutorial on {B}ayesian optimization.
\newblock \emph{arXiv preprint arXiv:1807.02811}, 2018.

\bibitem[Bubeck et~al.(2011)Bubeck, Munos, Stoltz, and
  Szepesv{\'a}ri]{bubeck2011x}
S{\'e}bastien Bubeck, R{\'e}mi Munos, Gilles Stoltz, and Csaba Szepesv{\'a}ri.
\newblock X-armed {B}andits.
\newblock \emph{Journal of Machine Learning Research}, 12\penalty0 (5), 2011.

\bibitem[Munos(2011)]{munos2011optimistic}
R{\'e}mi Munos.
\newblock Optimistic optimization of a deterministic function without the
  knowledge of its smoothness.
\newblock \emph{Advances in neural information processing systems},
  24:\penalty0 783--791, 2011.

\bibitem[Kleinberg et~al.(2008)Kleinberg, Slivkins, and
  Upfal]{kleinberg2008multi}
Robert Kleinberg, Aleksandrs Slivkins, and Eli Upfal.
\newblock Multi-armed bandits in metric spaces.
\newblock In \emph{Proceedings of the fortieth annual ACM symposium on Theory
  of computing}, pages 681--690, 2008.

\bibitem[Valko et~al.(2013)Valko, Carpentier, and Munos]{valko2013stochastic}
Michal Valko, Alexandra Carpentier, and R{\'e}mi Munos.
\newblock Stochastic simultaneous optimistic optimization.
\newblock In \emph{International Conference on Machine Learning}, pages 19--27.
  PMLR, 2013.

\bibitem[Pisier(1999)]{pisier1999volume}
Gilles Pisier.
\newblock \emph{The volume of convex bodies and Banach space geometry},
  volume~94.
\newblock Cambridge University Press, 1999.

\bibitem[Grosse et~al.(2021)Grosse, Zhang, and Hennig]{grosse2021probabilistic}
Julia Grosse, Cheng Zhang, and Philipp Hennig.
\newblock Probabilistic {DAG} search.
\newblock In \emph{Uncertainty in Artificial Intelligence}, pages 1424--1433.
  PMLR, 2021.

\bibitem[Rando et~al.(2022)Rando, Carratino, Villa, and Rosasco]{rando2021ada}
Marco Rando, Luigi Carratino, Silvia Villa, and Lorenzo Rosasco.
\newblock {Ada-BKB}: {S}calable {G}aussian {P}rocess {O}ptimization on
  {C}ontinuous {D}omains by {A}daptive {D}iscretization.
\newblock In \emph{International Conference on Artificial Intelligence and
  Statistics}, pages 7320--7348. PMLR, 2022.

\bibitem[Talagrand(1996)]{talagrand1996majorizing}
Michel Talagrand.
\newblock Majorizing measures: the generic chaining.
\newblock \emph{The Annals of Probability}, 24\penalty0 (3):\penalty0
  1049--1103, 1996.

\bibitem[Borst et~al.(2020)Borst, Dadush, Olver, and
  Sinha]{borst2020majorizing}
Sander Borst, Daniel Dadush, Neil Olver, and Makrand Sinha.
\newblock Majorizing measures for the optimizer.
\newblock \emph{arXiv preprint arXiv:2012.13306}, 2020.

\bibitem[Talagrand(2021)]{talagrand2021empirical}
Michel Talagrand.
\newblock Empirical processes, ii.
\newblock In \emph{{U}pper and {L}ower {B}ounds for {S}tochastic {P}rocesses},
  pages 433--456. Springer, 2021.

\bibitem[Shekhar and Javidi(2018)]{shekhar2018gaussian}
Shubhanshu Shekhar and Tara Javidi.
\newblock {G}aussian process bandits with adaptive discretization.
\newblock \emph{Electronic Journal of Statistics}, 12\penalty0 (2):\penalty0
  3829--3874, 2018.

\bibitem[Gonzalez(1985)]{gonzalez1985clustering}
Teofilo~F Gonzalez.
\newblock Clustering to minimize the maximum intercluster distance.
\newblock \emph{Theoretical computer science}, 38:\penalty0 293--306, 1985.

\bibitem[Chaslot et~al.(2008)Chaslot, Bakkes, Szita, and
  Spronck]{chaslot2008monte}
Guillaume Chaslot, Sander Bakkes, Istvan Szita, and Pieter Spronck.
\newblock Monte-{C}arlo tree search: A new framework for game ai.
\newblock In \emph{Proceedings of the AAAI Conference on Artificial
  Intelligence and Interactive Digital Entertainment}, volume~4, pages
  216--217, 2008.

\bibitem[Grill et~al.(2018)Grill, Valko, and Munos]{grill2018optimistic}
Jean-Bastien Grill, Michal Valko, and R{\'e}mi Munos.
\newblock Optimistic optimization of a {B}rownian.
\newblock \emph{Advances in Neural Information Processing Systems}, 31, 2018.

\bibitem[Wang et~al.(2014)Wang, Shakibi, Jin, and Freitas]{wang2014bayesian}
Ziyu Wang, Babak Shakibi, Lin Jin, and Nando Freitas.
\newblock Bayesian multi-scale optimistic optimization.
\newblock In \emph{Artificial Intelligence and Statistics}, pages 1005--1014.
  PMLR, 2014.

\bibitem[Gupta et~al.(2021)Gupta, Rana, Venkatesh, et~al.]{gupta2021bayesian}
Sunil Gupta, Santu Rana, Svetha Venkatesh, et~al.
\newblock {B}ayesian {O}ptimistic {O}ptimisation with {E}xponentially
  {D}ecaying {R}egret.
\newblock In \emph{International Conference on Machine Learning}, pages
  10390--10400. PMLR, 2021.

\bibitem[Contal et~al.(2015)Contal, Malherbe, and
  Vayatis]{contal2015optimization}
Emile Contal, C{\'e}dric Malherbe, and Nicolas Vayatis.
\newblock Optimization for {G}aussian processes via chaining.
\newblock \emph{arXiv preprint arXiv:1510.05576}, 2015.

\bibitem[Surjanovic and Bingham(2022)]{simulationlib}
S.~Surjanovic and D.~Bingham.
\newblock Virtual library of simulation experiments: Test functions and
  datasets.
\newblock Retrieved May 16, 2022, from \url{http://www.sfu.ca/~ssurjano}, 2022.

\bibitem[Paleyes et~al.(2019)Paleyes, Pullin, Mahsereci, Lawrence, and
  González]{emukit2019}
Andrei Paleyes, Mark Pullin, Maren Mahsereci, Neil Lawrence, and Javier
  González.
\newblock Emulation of physical processes with emukit.
\newblock In \emph{Second Workshop on Machine Learning and the Physical
  Sciences, NeurIPS}, 2019.

\bibitem[Gardner et~al.(2018)Gardner, Pleiss, Bindel, Weinberger, and
  Wilson]{Gardner2018}
Jacob~R Gardner, Geoff Pleiss, David Bindel, Kilian~Q Weinberger, and
  Andrew~Gordon Wilson.
\newblock {GPyTorch}: {B}lackbox matrix-matrix {G}aussian process inference
  with {GPU} acceleration.
\newblock \emph{Advances in Neural Information Processing Systems (NeurIPS)},
  2018:\penalty0 7576--7586, 2018.

\end{thebibliography}
\newpage
\subsection{Theoretical Analysis}

\subsubsection{Background}

The following three facts will be used during the analysis:

\begin{enumerate}
    \item \textbf{Deviation inequality} For a sample $f \sim \mathcal{G}(0,k)$ from a centered GP with kernel function $k$, it holds
    \begin{align}
    \label{eqn:gpdeviation}
        \forall u, \mathbb{P}(|f(x)-f(y)| \geq u) \leq 2 \exp \biggl( - \frac{u^2}{2d(x,y)^2} \biggr),
    \end{align}
    where $d(x,y) = \sqrt{k(x,x)+k(y,y)-2k(x,y)}$.
    \item \textbf{Hölder's inequality} 
    Let $a_1,...,a_N, b_1, ..., b_N$ be real numbers.
    \begin{align}
     \label{eqn:hölderineuality}
        \sum_{n=1}^N |a_nb_n| \leq \biggl(\sum_{n=1}^N |a_n|^p \biggr)^{1/p}  \biggl(\sum_{n=1}^N |b_k|^q \biggr)^{1/q}
    \end{align}
    with $p,q \geq 1$ and $\frac{1}{p}+\frac{1}{q}=1$
    \item \textbf{Growth of Harmonic numbers}
    The $N$-th Harmonic number $H_N = \sum_{n=1}^N \frac{1}{n}$ grows logarithmically in $N$:
    \begin{align}
    \label{eqn:harmonicnumber}
    H_N \in \Theta(\log N)
    \end{align}
\end{enumerate}

\subsubsection{Upper bounds on the supremum of a cell}
\textbf{Lemma 1.}
Assume $\mathcal{X}$ is finite and $f \sim \mathcal{GP}(0,k)$. Pick $\epsilon \in (0,1)$ and set $\beta_n =2\log(2|\mathcal{X}_n|N /\epsilon)$, where $\sum_{n \geq 1} \pi_n^{-1} =1, \pi_n > 0$. Then
\begin{align*}
    \forall n \sup_{x \in \mathcal{X}_n} |f(x)-f(x_n)| \leq \beta_n^{1/2}\Delta(\mathcal{X}_n)
\end{align*}
holds with probability $\geq 1-\epsilon$. $\mathcal{X}_n$ is the cell visited at step $n$ with center $x_n$ and $\Delta(\mathcal{X}_n) = \sup_{x \in \mathcal{X}_n} d(x,x_n)$. 

\textbf{Proof of Lemma 1.}
Fix $n$. Applying a union bound and \ref{eqn:gpdeviation}, one obtains for all $u_n$
\begin{align}
\mathbb{P}(sup_{x \in \mathcal{X}_n} |f(x)-f(x_{n})| \geq u_n)\\
 \leq \sum_{x \in \mathcal{X}_n} \mathbb{P}(|f(x)-f(x_{n})|\geq u_n)\\
 \leq 2 \sum_{x \in \mathcal{X}_n} \exp \biggl(- \frac{u_n^2}{2d(x,x_n)^2} \biggr)\\
 \leq 2 |\mathcal{X}_n| \exp \biggl(-\frac{u_n^2}{2\Delta(\mathcal{X}_n)^2}\biggr)
\end{align}\\
For $u_n = \beta_n^{1/2}\Delta(\mathcal{X}_n)$, it holds with probability $1-\epsilon/ N$, that $sup_{x \in \mathcal{X}_n} |f(x)-f(x_{n)})| \leq \beta_n^{1/2}\Delta(\mathcal{X}_n)$. Taking another union bound over $N$ the statement holds.

\subsubsection{Upper bound on the regret}
\textbf{Lemma 2.}
Running GP-OO with $\beta_n$ as specified in Lemma 1 and the canonical pseudo-metric $d$, the simple regret $r_n=f^*-f_n$ is bounded by $\beta_n^{1/2}\Delta(\mathcal{X}_n)$ for all $n$ with probability $1- \epsilon$.

\textbf{Proof of Lemma 2.}
This statement can be shown with typical arguments from the literature on optimistic optimization.
We consider the cases $x^* \in \mathcal{X}_n$ and $x^* \notin \mathcal{X}_n$ separately. \\

Case 1: $x^* \in \mathcal{X}_n$
\begin{align}
r_n = f^*-f_n = sup_{x \in \mathcal{X}_n} |f(x) -f(x_n)| \leq \beta_n^{1/n} \Delta(\mathcal{X}_n) 
\end{align}
by Lemma 1.\\

Case 2: $x^* \notin \mathcal{X}_n$.\\
Because $x_n$ was explored nevertheless, there is a node $x_{n'}$ on the optimal path with $x^* \in \mathcal{X}_{n'}$ , that was explored at step $n'<n$, such that
\begin{align}
    f(x_n) + \beta_n^{1/2} \Delta(\mathcal{X}_n)\geq f(x_{n'}) + \beta_{n'}^{1/2}\Delta(\mathcal{X}_{n'})
\end{align}
Then, $f(x_{n'}) - f(x_n) \leq \beta_n^{1/2} \Delta(\mathcal{X}_n)- \beta_{n'}^{1/2}\Delta(\mathcal{X}_{n'})$. For the regret one obtains in combination with Lemma 1:
\begin{align}
    r_n = f^*-f_n\\
    \leq \bigl[f(x^*)-f(x_{n'})\bigr] + \bigl[f(x_{n'})-f(x_{n})\bigr] \\
    \leq \bigl[ \beta_{n'}^{1/2}\Delta(\mathcal{X}_n')\bigr] + \bigl[\beta_n^{1/2} \Delta(\mathcal{X}_n)- \beta_{n'}^{1/2}\Delta(\mathcal{X}_{n'}) \bigr]\\
    = \beta_n^{1/2} \Delta(\mathcal{X}_n)
\end{align}

\textbf{Proposition 1.}
Assume $\mathcal{X}$ is finite and $f \sim \mathcal{GP}(0,k)$. Pick $\epsilon \in (0,1)$ and set $\beta_n =2\log(2|\mathcal{X}_n|N/\epsilon)$. For GP-OO with $k=2$ children nodes, we obtain the following bound on the cumulative regret
\begin{align}
\mathbb{P}(R_N \leq \beta^{1/2} \sum_{n=1}^N\Delta(\lfloor \log n \rfloor)) \geq 1-\epsilon
\end{align}
where $\Delta(h)$ is the radius of a cell at depth $h$ and $\beta = \max\{\beta_1, ..., \beta_N\}$.

\textbf{Proof Proposition 1.}
Simple consequence from Lemma 2, where we assume the worst case of a uniformly growing tree. The depth of a node after $n$ steps is at least $\lfloor \log(n) \rfloor$ in a uniformly growing tree.

\subsubsection{Bounds for common kernels}

The following analysis is restricted to GP's, where the canonical pseudo-metric $d$ satisfies:

\textbf{Assumption 1.}
There exist $C > 0, \alpha > 0$ such that $d(x,y) \leq C \|x-y\|_2^{\alpha}$, where $\|\cdot\|_2$ is the Euclidean norm. We additionally require $m/\alpha > 1$, where $m$ is the dimension of the domain.

According to \cite{shekhar2018gaussian} the first part of Assumption 1 holds for the squared exponential kernel with $C = \sqrt{2/l}, \alpha = 1$ and the Matern kernels with half integer values. For $\nu = 1/2$, one has $\alpha = 1/2$ and for all other half-integer values $\alpha=1$.

\textbf{Lemma 3.} [\cite{bubeck2011x}]
Assume that $\mathcal{X}$ is a $m$-dimensional hypercube $[0,1]^m$ and consider the dissimilarity $d(x,y) = C \|x-y\|_2^{\alpha}$, where $C > 0, \alpha > 0$. Define the partitioning by recursively splitting the hypercube in the middle along its longest side (ties broken arbitrarily). One has
\begin{align*}
    \Delta(\mathcal{X}_n) \leq diam(\mathcal{X}_{h}) \leq C(2 \sqrt{m})^{\alpha} \biggl(\frac{1}{2^{\alpha/m}}\biggr)^h
\end{align*}
for the cell $\mathcal{X}_{h}$ of a node at depth $h$.\\

\textbf{Proof of Lemma 3.} See Example 1 in \cite{bubeck2011x}.

\textbf{Proposition 2}
\textit{Assume the GP's canonical pseudo-metric $d$ satisfies $d(x,y) = C\|x-y\|_2^{\alpha}$, where $C > 0, \alpha > 0, m/\alpha > 1$. Running GP-OO on a finite domain $ \mathcal{X} \subset [0,1]^m$ with regular partitions, one has a worst-case cumulative regret $R_N$ of}
\begin{align*}
\tilde{\mathcal{O}}(N^{1-\alpha/m}(\log N)^{\alpha/m})
\end{align*}
\textit{with high probability. The $\tilde{\mathcal{O}}(\cdot)$ notation supresses poly-logarithmic factors}.\\

\textbf{Proof of Proposition 2}
It follows from Proposition 1 that $R_N \in \tilde{\mathcal{O}}(\sum_{n=1}^N\Delta(\lfloor \log n \rfloor))$ with probability $1-\epsilon$. It remains to bound $\sum_{n=1}^N \Delta(\lfloor\log n \rfloor)$. For Equation (\ref{eqn:lemma3}), we use Lemma 3 and for Equation (\ref{eqn:hölderineualityappl}), we apply Hölder's inequality (\ref{eqn:hölderineuality}) with $q=m/\alpha$ and $p=\frac{1}{1-\alpha/m}$.
\begin{align}
    \label{eqn:lemma3}
    \sum_{n=1}^N \Delta(\lfloor\log n \rfloor) \leq C \bigl(2 \sqrt{m} \bigr)^{\alpha} \sum_{n=1}^N \biggl(\frac{1}{2^{\alpha/m}} \biggr)^{\lfloor \log n \rfloor}\\
    \leq C(2 \sqrt{m})^{\alpha} \sum_{n=1}^N \biggl(\frac{1}{2^{\alpha/m}} \biggr)^{\log n - 1}\\
    = C(2 \sqrt{m})^{\alpha}2^{\alpha/m}\sum_{n=1}^N \biggl(\frac{1}{2^{\alpha/m}} \biggr)^{\log n}\\
    = C_1 \sum_{n=1}^{N} \biggl(\frac{1}{2^{\alpha/m}} \biggr)^{\log n}\\
    = C_1 \sum_{n=1}^{N} \biggl(\frac{1}{n^{\alpha/m}}\biggr)\\
    \label{eqn:hölderineualityappl}
    \leq C_1 \biggl( \sum_{n=1}^N 1\biggr)^{1-\alpha/m} \biggl(\sum_{n=1}^N \biggl(\frac{1}{n^{\alpha/m}}\biggr)^{m/\alpha} \biggr)^{\alpha/m}\\
    = C_1 N^{1-\alpha/m} H_N^{\alpha/m}
\end{align}
where $C_1 = C(2 \sqrt{m})^{\alpha}2^{\alpha/m}$ and $H_N$ being the $N$-th harmonic number. Together with (\ref{eqn:harmonicnumber}), this implies:
\begin{align}
    \sum_{n=1}^N \Delta(\lfloor\log n \rfloor) \in \mathcal{O}(N^{1-\alpha/m}(\log N)^{\alpha/m})
\end{align}

\subsection{Additional experimental details}

All experiments were implemented in Python 3.9.1. and run on a machine with macOS 12.3.1, a 4 GHz Quad-Core Intel Core i7 CPU and 32 GB RAM. \\
\subsubsection{Experiment with benchmark functions}
\begin{figure}
    \centering
    \includegraphics{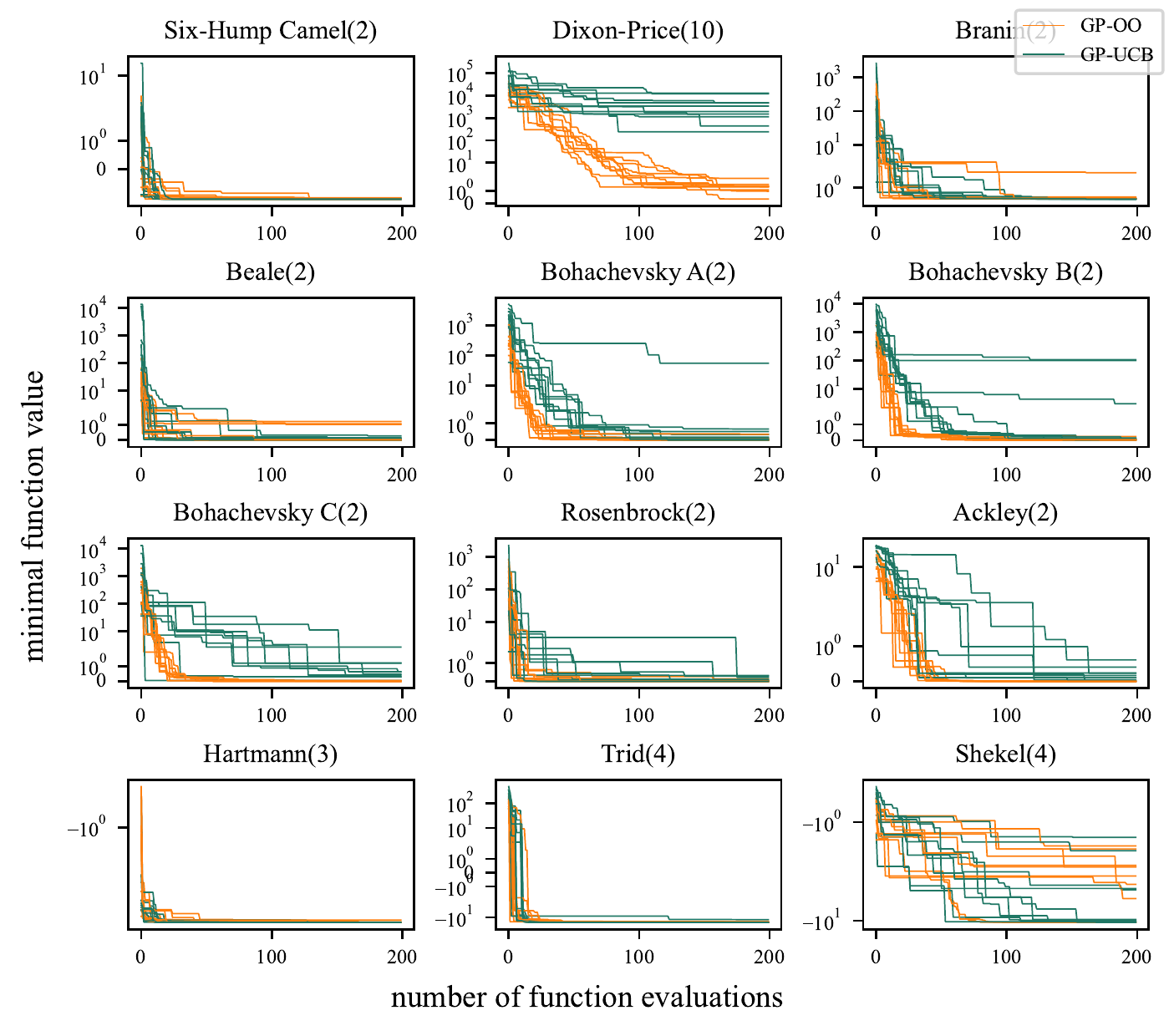}
    \caption{Minimal function values found on common benchmark datasets.}
    \label{fig:benchmark_regret}
\end{figure}

Table \ref{sample-table} lists the domains and hyperparameters used in the experiment with the benchmark functions. For each run, we sampled a sub-domain by choosing uniformly at random new lower and upper boundaries for the intervals along each dimension, such that the new boundaries are inbetween the previous ones and the location of the minimum. In this way, the location of the minimum stays the same as in the original domain. Figure \ref{fig:benchmark_regret} shows the minimal function value found for the benchmark functions.
\begin{table}
  \caption{Hyperparameters for experiment with benchmark functions}
  \label{sample-table}
  \centering
  \begin{tabular}{lllll}
    \toprule
    Benchmark    & Domain     & Lengthscale $l$ & $\beta$ (GP-UCB) & $\beta$ (GP-OO) \\
    \midrule
    Branin & $[-15,15]^2$  & $0.5$ &  $10$  & $100$\\
    Six-Hump-Camel     & $[-2,2]^2$& $0.5$  &  $1$ & $100$ \\
    Beale     & $[-4.5, 4.5]^2$       &$1$ & $1$& $100$\\
    Bohachevsky a    & $[-35.5, 100]^2$    &  $1.7$  & $10$& $10$\\
    Bohachevsky b    & $[-35.5, 100]^2$        & $1.7$ & $10$ & $10$\\
    Bohachevsky c    & $[-35.5, 100]^2$        & $1.7$ & $100$& $10$\\ 
    Rosenbrock   & $[-3,3]^2$      & $0.7$  & $1$& $100$\\
    Ackley   & $[-12.5,35]^2$       & $3.5$  & $10$ & $10$\\
    Hartmann   & $[0,1]^3$      & $0.3$  & $1$ & $10$\\
    Trid   & $[-16, 16]^4$      & $10.75$  & $0.1$ & $100$\\
    Shekel  & $[0, 10]^4$       & $1.75$  & $10$ & $10$\\
    Dixonprice & $[-10,10]^{10}$& $2$&  $1$ & $10$\\
    \bottomrule
  \end{tabular}
\end{table}

\end{document}